# Enhancing Abstractive Summarization of Scientific Papers Using Structure Information

Tong Bao, Heng Zhang, Chengzhi Zhang*

Department of Information Management, Nanjing University of Science and Technology, Nanjing, 210094, China
{tbao, zh_heng, zhangcz}@njust.edu.cn

**Abstract:** Abstractive summarization of scientific papers has always been a research focus. yet existing methods face two main challenges. First, most summarization models rely on Encoder-Decoder architectures that treat papers as sequences of words, thus fail to fully capture the structured information inherent in scientific papers. Second, existing research often use keyword mapping or feature engineering to identify the structural information, but these methods struggle with the structural flexibility of scientific papers and lack robustness across different disciplines. To address these challenges, we propose a two-stage abstractive summarization framework that leverages automatic recognition of structural functions within scientific papers. In the first stage, we standardize chapter titles from numerous scientific papers and construct a large-scale dataset for structural function recognition. A classifier is then trained to automatically identify the key structural components (e.g., *Background*, *Methods*, *Results*, *Discussion*), which provides a foundation for generating more balanced summaries. In the second stage, we employ Longformer to capture rich contextual relationships across sections and generating context-aware summaries. Experiments conducted on two domain-specific scientific paper summarization datasets demonstrate that our method outperforms advanced baselines, and generates more comprehensive summaries.[1]

---

# 1. Introduction

With the rapid growth of scientific research and the academic community, numerous scientific papers are published daily. This notable increase in publications has led to information overload and requiring scholars to spend considerable time in reading and comprehending a large volume of articles. The goal of automatic summarization is to employ algorithms to extract key information and reorganize it into shorter, concise summaries (El-Kassas et al., 2021). Automatic summarization holds significant research value in fields such as information retrieval (Spina et al., 2017), question and answer system (Yulianti et al., 2018), and content review(Hu et al., 2017). Existing automatic summarization methods are broadly divided into two categories: extractive methods and abstractive methods. Extractive methods generate summaries by selecting sentences directly from the original document, resulting in summaries that are more accurate and semantically consistent but may lack coherence. In contrast, abstractive methods generate summaries based on an understanding of the text, rather than extracting sentences directly from the original document. Therefore, summaries produced by this approach are typically more coherent and better aligned with human reading preferences (El-Kassas et al., 2021; Ghadimi & Beigy, 2022). In this paper, we focus on abstractive summarization.

**Exploratory Biomarker Analysis Using Plasma Angiogenesis-Related Factors and Cell-Free DNA in the TRUSTY Study: A Randomized, Phase II/III Study of Trifluridine/Tipiracil Plus Bevacizumab as Second-Line Treatment for Metastatic Colorectal Cancer**

Yu Sunakawa[1] · Yasutoshi Kuboki[2] · Jun Watanabe[3] · Tetsuji Terazawa[4] · Hisato Kawakami[5] · Mitsuru Yokota[6] · Masato Nakamura[7] · Masahito Kotaka[8] · Naotoshi Sugimoto[9] · Hitoshi Ojima[10] · Eiji Oki[11] · Takeshi Kajiwara[12] · Yoshiyuki Yamamoto[13] · Yasushi Tsuji[14] · Tadamichi Denda[15] · Takao Tamura[16] · Soichiro Ishihara[17] · Hiroya Taniguchi[18] · Takako Eguchi Nakajima[19] · Satoshi Morita[20] · Kuniaki Shirao[21] · Naruhito Takenaka[22] · Daisuke Ozawa[22] · Takayuki Yoshino[23]

**Abstract**

**Background** The TRUSTY study evaluated the efficacy of second-line trifluridine/tipiracil (FTD/TPI) plus bevacizumab in metastatic colorectal cancer (mCRC).

**Objective** This exploratory biomarker analysis of TRUSTY investigated the relationship between baseline plasma concentrations of angiogenesis-related factors and cell-free DNA (cfDNA), and the efficacy of FTD/TPI plus bevacizumab in patients with mCRC.

**Patients and Methods** The disease control rate (DCR) and progression-free survival (PFS) were compared between baseline plasma samples of patients with high and low plasma concentrations (based on the median value) of angiogenesis-related factors. Correlations between cfDNA concentrations and PFS were assessed.

**Results** Baseline characteristics (*n* = 65) were as follows: male/female, 35/30; median age, 64 (range 25–84) years; and *RAS* status wild-type/mutant, 29/36. Patients in the hepatocyte growth factor (HGF)-low and interleukin (IL)-8-low groups had a significantly higher DCR (risk ratio [95% confidence intervals {CIs}]) than patients in the HGF-high [1.83 [1.12–2.98]] and IL-8-high (1.70 [1.02–2.82]) groups. PFS (hazard ratio [HR] [95% CI]) was significantly longer in patients in the HGF-low (0.33 [0.14–0.79]), IL-8-low (0.31 [0.14–0.70]), IL-6-low (0.19 [0.07–0.50]), osteopontin-low (0.39 [0.17–0.88]), thrombospondin-2-low (0.42 [0.18–0.98]), and tissue inhibitor of metalloproteinase-1-low (0.26 [0.10–0.67]) groups versus those having corresponding high plasma concentrations of these angiogenesis-related factors. No correlation was observed between cfDNA concentration and PFS.

**Conclusion** Low baseline plasma concentrations of HGF and IL-8 may predict better DCR and PFS in patients with mCRC receiving FTD/TPI plus bevacizumab, however further studies are warranted.

**Clinical Trial Registration Number** jRCTs031180122.

(a) Structural information within the structured abstract.

(b) Structural information within the flat abstract

Figure 1. Examples of structural information in both structured and flat abstracts of scientific papers. The source article is from the paper (Sunakawa et al., 2024) and (Fu et al., 2020).

Despite sequence-to-sequence (seq2seq) models have achieved impressive results on relatively short documents such as news and user comments (Nallapati et al., 2016), scientific articles differ from these in several key aspects. First, the average length of scientific papers exceeds 3,000 words, while news articles typically average around 700 words(Gidiotis & Tsoumakas, 2020). This increased length raises computational complexity and introduces challenges in handling long-distance dependencies for generative models. Secondly, scientific papers follow a structured format known as IMRaD (*Introduction, Methods, Results,* and *Discussion*)(Andrade, 2011). Each section serves a specific purpose, and the summarization process need consider the hierarchical structure and internal connections between these sections (Figure 1). Lastly, news article summaries typically cover key information into about 100 words, a scientific paper summary needs to comprehensively summarize content from multiple sections, often exceeding 200 words and sometimes reaching up to 400 words(Oh et al., 2023). These differences pose challenges that current methods unsuitable for direct application to automatic summarization of scientific papers.

Existing methods to improving scientific paper summarization primarily focus on two research directions. The first direction aims to extracting key information to reduce the length of papers. Techniques such as topic models or heterogeneous trees are commonly used by researchers to selectively extract important words or sentence from the text. For example, Zhu et al. (2023) employed a heterogeneous tree structure and triplet position to perform extractive summarization of scientific papers. Additionally, Han, Feng, and Qi (2024) introduced TopicSum, a framework that uses a heterogeneous graph neural network to leverage topic information as document-level features for sentence selection. Despite their effort, these methods often overlook semantic and logical consistency between the generated summaries and the original text, leading to incomplete and shallow summaries. Another research direction focuses on improve the efficiency of summarizing long texts. Models such as Longformer (Beltagy et al., 2020) and BigBird (Zaheer et al., 2020) have been proposed for this purpose. However, the semantically connected nature of sections in scientific papers presents challenges for directly summarizing them into coherent abstracts. Some extractive summarization studies adopt divide-and-conquer approaches, extracting key sentences from each section and then connecting them into a summary (Gidiotis & Tsoumakas, 2020; Miao et al., 2019). However, these studies only use rules-based methods to locate the first-level titles of the sections, ignoring the fact that scientific papers do not follow a uniform structure, in other words, the titles provided by authors exhibit flexibility and diversity, which is particularly evident across different fields. For example, in the computer science field, when authors propose a new model, they tend to name the chapter according to their specific model(Zhang et al., 2024a). In such cases, it is difficult to accurately locate titles and corresponding content using rule-based methods. This limitation may lead to decline quality of the generated summaries, as the system fail to receive the necessary section content and the function information. Additionally, these studies treat the extraction of structure information and summary generation as separate tasks, requiring a re-extraction of structural information for different document structures or summarization contexts, thus overlook the potential benefits of integrating these two processes.

   To address this gap, we propose a two-stage abstractive summarization framework for scientific articles based on structural function recognition. First, we collected original articles from arXiv and PubMed, and standardized the section headings according to the IMRaD format to construct a large-scale dataset for structural function recognition. Second, a classifier was developed to identify structural function categories within the chapters, where the beginning and ending portions of chapters, which contain the highest information density, were selected to improve classification performance. Finally, to fully grasp contextual information, the identified chapter labels, along with the corresponding content, were fed into Longformer, a model designed specifically for processing long documents, to generate summaries for scientific papers. We conducted thorough experiments on two public datasets and demonstrated the advantages of our approach in scientific paper abstractive summarization. The main contributions of this paper are as follows:

- We proposed a two-stage framework for abstractive summarization of scientific papers, which leverages automatic structural function recognition to generate more balanced and comprehensive summaries. To the best of our knowledge, we are the first to integrate automatic structural function recognition into the task of abstractive summarization for scientific papers.

- A controlled experiment demonstrated that the beginning and ending positions of chapters contain important information, which is beneficial for structural function recognition. Additionally, through systematic evaluation, we pointed out the biases in traditional summary quality metrics when evaluating generative models, such as GPT-4.

- A large-scale dataset was constructed to recognize structural functions in scientific papers.

- Experiments on two benchmarks demonstrate that our method outperforms advanced baselines in scientific paper abstractive summarization, with the generated summaries being more comprehensive than those from the baseline models.

The rest of the paper is structured as follows: Section 2 summarizes related works.

We present our method in Section 3, and the experiments are provided in Section 4. The results are reported in Section 5. We discuss the implications and limitations of this paper in Section 6. Finally, the conclusion and future work are outlined in Section 7.

## 2. Related work

In this section, we first review related works on abstractive summarization, specifically focus on summarization in scientific papers. We then briefly introduce research on chapter structure recognition within scientific papers.

### 2.1 Abstractive summarization

Abstractive Summarization aims to generate summaries by understanding the meaning of the text, rather than simply extracting key sentences or phrases from the original documents. Khan(2014), Mohan et al.(2016) and Oh et al.(2023) have classified existing abstractive summarization methods into three categories: structure-based approaches(Ganesan et al., 2010; L. Wang & Ling, 2016), semantic-based approaches(Alshaina et al., 2017; Khan et al., 2018; Mohan et al., 2016) and deep learning approaches. Over the past decade, the seq2seq architecture based on Recurrent Neural Networks (RNNs) has achieved widespread success in NLP tasks, making deep learning approaches the dominant framework for abstractive summarization (Hou et al., 2018). For instance, Rush et al. (2015) were the first to apply the encoder-decoder framework in automatic text summarization tasks. Gu et al.(2016) introduced a coverage mechanism to addresses out-of-vocabulary (OOV) issues by copying words directly from the original text with a certain probability. Cohan et al. (2018) proposed a hierarchical decoder structure that treats the documents as a collection of paragraphs, achieving strong performance on summarization tasks involving lengthy articles.

With the advent of pre-trained language models, researchers have started shifting from traditional deep learning approaches to leveraging pre-training models in pursuit of better summarization performance. Liu and Lapata(2019) proposed BERTSUM, which applied the pre-training BERT (Devlin et al., 2019) model to abstractive summarization. Aksenov et al. (2020) optimized BERT's encoding process using a sliding window strategy, improving its effectiveness for longer documents. Liu, Cao, and Yang (2022) incorporated phrase-level prior knowledge into the Transformer-based summarization model to improve summarization quality. Following this work, researchers have further enhanced the generative summarization by advancing the BERT-based architectures, such as PEGASUS(Zhang et al., 2020), T5(Raffel et al., 2020), BART(Lewis et al., 2019) and BigBird(Zaheer et al., 2020). However, these methods encounter challenges when processing long documents or addressing OOV issues, which can lead to disjointed or repetitive summaries (Hou et al., 2018; Lund et al., 2023; Miao et al., 2019;). Recently, the emergence of LLMs, such as ChatGPT, has significantly reshaped the paradigms of NLP tasks. Several studies have demonstrated that LLMs surpass human-level performance in abstractive summarization tasks. For example, Basyal and Sanghvi (2023) conducted a comparative study exploring the performance of various LLMs on text summarization, revealing the broad potential of LLMs in cross-domain summarization. Deroy et al. (2024) applied general-domain LLMs in legal case judgments summarization and found that generative models produce summaries that are more aligned with human preferences. Zhang et al. (2024b) found that LLMs achieve high-quality news summarization through instruction tuning, and their evaluations revealed that LLM-generated summaries are comparable to those written by humans.

In summary, abstractive summarization has remained a key research focus, progressing from early structure-based approaches to the current LLMs-based methods, which have achieved remarkable performance. Despite these advancements, most summarization research has focused on general domains. Applying these methods to scientific articles, however, still presents unique challenges due to the complex structure and length of scientific texts. Therefore, in this paper, we focus on generative summarization for scientific articles. To achieve this, we proposed a two-stage abstractive summarization framework, which leverages automatic structural function recognition to generate more balanced and informative summaries. Experiments on two widely used datasets validated the effectiveness of automatic structural function recognition in abstractive summarization for scientific papers.

## 2.2 Abstractive summarization of scientific papers

Most abstractive summarization methods primarily designed for general domains and news articles. However, scientific papers differ greatly from news articles in terms of text length, structure, terminology, and audience, making direct application of these methods to scientific paper summarization unsuitable (Gidiotis & Tsoumakas, 2020; Andrade, 2011; Oh et al., 2023). Saggion and Lapalme(2000) introduced the "selective analysis" method, which automatically generate informative abstracts for scientific articles by identifying their main topics. Ganesan et al. (2010) presented a graph neural network-based summarization model that effectively generates concise, abstractive summaries of highly redundant opinions. Several studies have incorporated citation contexts in scientific paper summarization. Elkiss et al. (2008) generated abstracts by extracting key sentences from the citations of the target paper, while Galgani et al. (2015) combined citation themes, the target paper, and citation content to create comprehensive summaries. Elizalde et al. (2016) improved summaries by utilizing citations directed at the target paper. Lauscher et al. (2017) developed an automated abstracting system for scientific papers by ranking and categorizing key sentences from target papers. Agrawal et al. (2019) employed a K-Nearest Neighbors (KNN) classifier to extract elements such as the title, citation sentences, and abstract, to generating a more comprehensive summary.

Several researchers have effectively incorporated the unique features of scientific papers, such as graphical information and chapter length, to enhance summarization. Bhatia and Mitra (2012) applied traditional machine learning methods, manually creating a set of features from scientific papers to summarize document elements. Yang et al. (2016) developed a system using a data-weighted reconstruction approach to generate extended summaries that capture the most important aspects of a scientific article. He et al. (2016) proposed applying topic models to distinguish unique features for the effective generation of group-specific abstracts. Additionally, Erera et al.(2019) summarized computer science articles by processing user queries and key entities. Wang et al. (2023) selected impactful sentences and reordered them to generate abstracts for multiple scientific articles on a given topic. Luo et al. (2023) proposed a citation graph-based summarization framework, which integrates key contents from references to enhance summary relevance. Van Veen et al. (2024) employed mainstream LLMs for summarizing clinical medical texts, and achieved performance that surpassed that of medical experts.

The work most similar to ours is by Oh et al. (2023), who propose a divide-and-conquer method using full-text section information for structured abstract summarization. However, their approach relies on simple keyword mapping to identify chapter labels and trains separate summarization models for each section due to input length limitations. This restricts the model's flexibility in handling diverse chapter structures. Unlike their approach, we train a classifier on a large-scale chapter structure classifications corpus to automatically identify the structure function in scientific papers. Furthermore, for the abstractive summarization stage, we adopt Longformer as the backbone, which is better suited to capture longer contextual information, resulting in more comprehensive summaries compared to training separate summarization models for each section.

## 2.3 Structural function recognition within Scientific papers

The structural organization of a scientific paper not only offers readers a logical s for comprehension but also serves as the foundation for fine-grained knowledge analysis. The IMRaD format is widely adopted in the academic community as a standard for scientific writing and structural organization(Ma et al., 2022; Oh et al., 2023). Generally, automated recognition of the structural functions in scientific articles primarily includes two categories: rule-based methods and deep learning-based methods. For instance, Lin et al. (2006) leveraged the IMRaD format to annotate biomedical abstracts, achieving notable success through the incorporation of bi-gram features and the Hidden Markov Model (HMM). Kiela et al. (2014) extracted bi-grams, lexical information, and sentence structure features to recognize chapter structures using a semi-supervised method. Similarly, Cox et al. (2018) applied verb features and utilized random forest and decision tree algorithms to automatically parse the structure of medical literature.

With the rapid advancement of deep learning models, there has been a growing research trend in utilizing neural network models for the structural functional recognition of scientific papers. Dasigi et al. (2017) were first to introduced a structure recognition model that incorporates attention mechanisms and Long Short-Term Memory networks (LSTM). Ma et al. (2022) proposed a deep learning-based classification model that integrated contextual information and relative positional features, resulting in improved classification accuracy. As one of the text classifications tasks, structural function recognition of scientific articles has yielded competitive performance. However, these studies utilize structural information for abstractive summarization, focusing on first-level section headings or simple keyword matching to define section structures (Cohan et al., 2018). Since the headings provided by authors are highly flexible and often inconsistent, the summaries generated by these models may fail to capture the complete context and logical relationships within paragraphs, leading to incomplete and unbalanced summaries(Gidiotis & Tsoumakas, 2020; Miao et al., 2019).

To address this issue, we reconstruct a large-scale dataset for structural function recognition in scientific papers. Specifically, we collect full-text scientific papers from arXiv and PubMed datasets and further normalize all chapter headings into four commonly used categories: *Background*, *Methods*, *Result* and *Conclusion*. Based on this dataset, we employ the SciBERT model to train a classifier that automatically identifies the structural functions within the paper and achieves highly competitive classification results. The identified chapter functions and corresponding content are then fed into Longformer, which is a model specifically designed to handle long documents, to generate balanced summaries. Our framework integrates automatic structural function recognition with abstractive summarization to ensure that the generated summaries contain key information while maintaining the logic and comprehensiveness for scientific papers.

## 3. Methodology

In this section, we first introduce the overall framework of this study, and then provide a detailed explanation of the implementation process for our proposed method.

### 3.1 Overall framework of this study

As mentioned above, we propose a two-stage abstractive summarization method for scientific papers that leverages automatic structural function recognition. The overall workflow of the model illustrated in Figure 2.

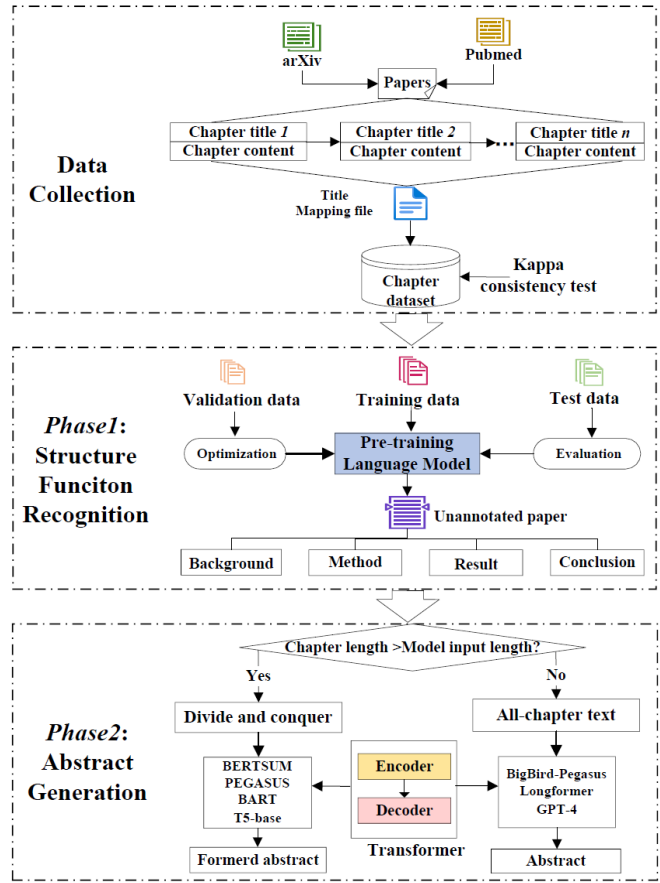

Figure 2．Overall framework of the study, where chapter function and their corresponding content are recognized and used as segmentation markers for the abstractive summarization phase.

The framework includes three components: data collection, structural function recognition, and abstract generation. First, we first collected chapter title and content pairs from papers in the arXiv and PubMed datasets. Then, these pairs were normalized using a title mapping file to construct a chapter structure classification dataset. During the structural function recognition phase, title-content pairs extracted from scientific papers are used to train the chapter structure recognition model, which classifies each unannotated chapter into predefined categories. The identified chapter labels, along their corresponding content, are then fed into the second abstract generation phase, where they serve as segmentation markers for the full-text article, guiding the summarization model to generate balanced and comprehensive summaries. For models with shorter input length limits than the combined length of all chapters, we apply a divide-and-conquer approach by generating individual summaries for each chapter and then concatenating them into the final summary. For models capable of processing longer documents, we input the full content of all chapters along with their function labels to produce a comprehensive summary.

## 3.2 Task formalization

The specific descriptions of the structural function recognition and the abstractive summarization are as follows:

**Structural function recognition (SFR):** Given a source document $\boldsymbol{D}$, $\{c_1, c_2, \ldots, c_n\}$ represents each chapter of $\boldsymbol{D}$, $n$ represents the number of structures in the given document, and a corresponding set of labels L = $\{l_1, l_2, \ldots, l_n\}$ represents the function categories of structures, the goal of SFR is to build a classification model $f$ that assigns each chapter $c_i$ to one predefined categories $l_i$.

**Abstractive Summarization (AS):** Given a source document $\boldsymbol{D}$, represented as a sequence of sentences $\{s_1, s_2, \ldots, s_n\}$, the objective of abstractive summarization is to generate a short and precise summary $\boldsymbol{S}$ that captures the key information and main ideas of $\boldsymbol{D}$. The summary is not constrained to be an extractive subset of sentences from $\boldsymbol{D}$，Unlike extractive methods, the summary is not limited to selecting sentences directly from the source text. Instead, it involves creating a new summary by rephrasing, paraphrasing, and synthesizing information in a more innovative and cohesive manner.

## 3.3 Model implementation

Our proposed abstractive summarization model includes two tasks: structural function recognition and abstractive summarization. Figure 3 illustrates the specific implementation process between these two tasks. The process begins with an input text that embedded with positional, segment, and token embeddings based on SciBERT.

These embeddings are then processed through multiple transformer layers to capture contextual information from the input text. The output from the transformer layers was fed into a Softmax function to assign a function label to each input text. At this point, we can automatically obtain the chapter structure information for each paper. Finally, an encoder-decoder architecture-based model is chosen to perform the abstractive summarization task, where the encoder generates a context vector from the classified text, and the decoder uses this vector to produce a summary for each identified section. In the following sections, we will provide the definition of each task and the implementation process.

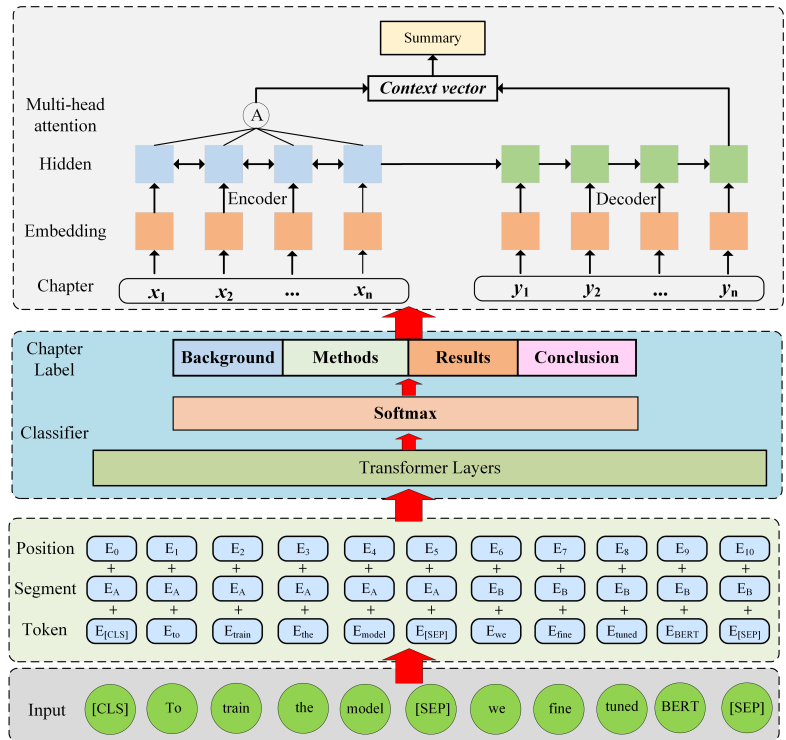

Figure 3. Abstractive summarization for scientific papers based on structural function recognition, where SciBERT and Longformer are selected as the backbones for two sequential tasks.

### 3.3.1 SFR model in scientific papers

In this task, to recognize the structure of scientific articles, we introduce a text classification model based on SciBERT, which is specifically pretrained on biomedical and computer science articles. Assuming the input chapter text is $C_n = (P_1, P_2, ..., P_n)$, where $C_n$ represents the chapter text composed of multiple paragraphs from $P_1$ to $P_n$. The chapter text $C_n$ undergoes sentence segmentation, resulting in a set of sentences $X = (X_1, X_2, ..., X_n)$. During this process, we employ the SciBERT model to vectorize each sentence, as shown in Figure 3.

In the encoding process, SciBERT introduces a [CLS] token at the beginning of the text as the starting annotation, and the final representation vector denoted as:

$$y_m = \text{Softmax}(H_{\text{CLS}}) \tag{1}$$

Next, the CLS vector is identified as the definitive feature representation vector of the chapter, serving as input for a fully connected layer. We then apply the Softmax function to obtain the final probability distribution over the label categories $m$.

$$y_m = \text{Softmax}(H_{\text{CLS}}) \tag{2}$$

Finally, we use the Cross-Entropy function to compute the loss between the predicted probability distribution $y_m$ and the true labels. This loss function is defined as:

$$Loss = -\frac{1}{m}\sum_{i=1}^{m}\sum_{j=1}^{n} y_{ij}\log\left(p(x_{ij})\right) \tag{3}$$

where $m$ is the number of samples, $n$ is the number of classes, $y_{ij}$ is the true label, which is 1 for the correct class and 0 for the others, and $p(x_{ij})$ is the predicted probability for class $j$ of the $i$-th sample.

### 3.3.2 AS model in scientific papers based on SFR

After completing the SFR task, we obtain the chapter contents and associated functional labels for each unlabeled article. These chapter content, along with their label marks, are then used as input text and segmentation labels for all generative models involved in the AS task to generate summaries. To obtain rich contextual information and accurately reflect the article's structure, we selected the Longformer model, which is specifically designed to process long documents, as the backbone of the summarization stage. Assuming the input chapter is represented as a sequence $X$，the input text is encoded through the Longformer Encoder layer to obtain embedded vectors $E$:

$$E = Longformer\_Encoder(X) \tag{4}$$

In the self-attention mechanism of Longformer, each embedding vector is transformed into three distinct vectors: Query (Q), Key (K), and Value (V)，to calculate attention scores and aggregate contextual information. The $Q$ vector represents the current token's focus or intent to attend to other tokens within the sequence. The $K$ vector encodes features of the tokens to be matched against the $Q$ vector, while the $V$ vector contains the actual information that will be weighted to produce the final output of the attention mechanism. The multi-layer self-attention in Longformer then computes relevance scores between these vector representations, which are used to generate contextually enriched token embeddings:

$$Attention\ (Q, K, V) = softmax\ \left(\frac{QK^T}{\sqrt{d_k}}\right) V \tag{5}$$

where $\sqrt{d_k}$ is used to normalize the dot product $QK^T$ to ensure the attention scores remain within a stable range. The *softmax* function is applied to converts the similarity scores into probabilities.

In the end, we obtain the corresponding output sentence of chapter through the Longformer Decoder layer.

$$Y = Longformer\_Decoder(E,\ Attention\ (Q, K, V)) \tag{6}$$

## 4. Experiments

We conduct comprehensive experiments to evaluate the performance of our proposed model. In this section, we will describe the details of the experiment, including the datasets, baselines, implementation details, and evaluation metrics.

### 4.1 Dataset

To facilitate comprehensive experiments and evaluate our proposed model, we restructured datasets for both the structural function recognition and abstractive summarization phases.

#### 4.1.1 Dataset of structural function recognition

We conduct experiments on two widely used scientific paper summarization datasets, PubMed and arXiv (Cohan et al., 2018), to evaluate the model's performance. The PubMed dataset consists of 133,215 articles from the biomedical domain, while the arXiv dataset comprises 215,913 articles from diverse scientific fields such as physics, computer science, and mathematics. Each article in both datasets provides metadata including article ID, abstract, section names, and section contents, but excluding figures and tables. To address the variability in chapter headings across scientific papers, we standardized the chapter titles according to the IMRaD format using National Library of Medicine (NLM) files(Guimarães, 2006). The NLM files contain 3032 unique chapter labels, categorized into five broader NLM categories: *Background*, *Objective*, *Method*, *Result*, and *Conclusion*. However, since this file was generated based on biomedical text and our dataset includes articles from arXiv, which spans multiple disciplines such as computer science, physics, and mathematics, we constructed a corpus based on these sources to train a classifier. As shown in Table 1, in this file, *Introduction* is categorized under the broader class of *Background*, while *Discussion* is grouped under *Conclusions*. Therefore, *Background*, *Method*, *Result*, and *Conclusion* represents a broader version of the IMRaD format. Notably, considering that the *Objective* is typically addressed in or following the *Background* section(Oh et al., 2023), we categorize it under *Background* category in this study.

Table 1．  Examples of NLM mapping for different chapter types.

| Type of Chapter | Example of NLM Categories |
|---|---|
| Background | Background, Introduction, Motivation, Hypothesis, Instruction, Aim |
| Method | Method, Methodology, Approach, Experiment, Measurement, Techniques |
| Result | Result, Finding, Evaluation, Innovations, Outcome, Output |
| Conclusion | Conclusion, Discussion, Impact, Implication, Summary, Limitation, Future work, Recommendation |

Subsequently, we randomly selected 30,000 articles from each dataset to construct the structural function recognition dataset. Table 2 presents the statistics of the selected papers from both PubMed and arXiv datasets.

Table 2．  Statistical for the structure function recognition datasets used in this work.

| Dataset \Chapter | PubMed | | | arXiv | | |
|---|---|---|---|---|---|---|
| | Num | $Word_{avg}$ | $Sent_{avg}$ | Num | $Word_{avg}$ | $Sent_{avg}$ |
| Background | 38678 | 390 | 14 | 35277 | 798 | 37 |
| Method | 80622 | 405 | 15 | 48164 | 989 | 44 |
| Result | 42746 | 549 | 20 | 34616 | 1425 | 67 |
| Conclusion | 61292 | 526 | 20 | 39535 | 858 | 40 |

*Note*: "Num" refers to the number of chapters in each category, "$Word_{avg}$" and "$Sent_{avg}$" represent the average number of words per chapter and the average number of sentences per chapter, respectively.

### *4.1.2 Dataset of abstractive summarization*

From the structural function recognition dataset, we additionally selected articles that contained the *Background*, *Methods*, *Results*, and *Conclusions* sections. Each section in both datasets was constrained to a maximum of 1500 words, and the abstract length ranged between 50 to 300 words. Finally, to account for economic costs and time constraints, we randomly sampled 10,000 papers from each dataset for abstractive summarization. Table 3 provides the statistics of these selected articles.

Table 3．Statistical features of the scientific article from arXiv and PubMed datasets.

| Dataset | #Doc | Avg.doc length | | Avg. abstract length | |
|---------|------|-------|-----------|-------|-----------|
| | | Words | Sentences | Words | Sentences |
| PubMed | 10K | 3107 | 92 | 201 | 8 |
| arXiv | 10K | 5042 | 217 | 225 | 10 |

## 4.2 Baselines

For the SFR task, we conduct a comparative analysis of various methods related to structural function recognition. The baselines include:

**(1) BiLSTM**[2](Dasigi et al., 2017): A widely-used sequence model in NLP tasks that captures contextual information from both preceding and succeeding directions in a sequence.

**(2) BERT**[3] (Devlin et al., 2019): One of the most successful pre-trained language models, often used as a standard baseline in automatic summarization tasks.

**(3) RoBERTa**[4](Liu et al., 2019): An enhanced version of BERT, trained on a larger dataset and improved training techniques.

**(4) T5-base**[5](Raff el al et al., 2020): A text-to-text Transformer model that classifies category labels in a generates manner.

For the AS task, we compare Longformer with several well-known PLMs. Notable models include:

---

[2] https://github.com/edvisees/sciDT

[3] https://huggingface.co/google-bert/bert-base-uncased

[4] https://huggingface.co/FacebookAI/roberta-base

[5] https://huggingface.co/google-t5/t5-base

**(1) BERTSUM**[6](Liu and Lapata, 2019): A model that modified the transformer architecture of BERT to text summarization tasks.

**(2) BART**[7](Lewis et al., 2019): A model based on Bidirectional and Auto-Regressive Transformers, excelling in various generative tasks through sequence-to-sequence pre-training.

**(3) PEGASUS**[8](Zhang et al., 2020): A large pre-trained model specifically designed for text summarization tasks.

**(4) T5-base** (Raffel et al., 2020): A unified text-to-text Transformer model treating all tasks as generative tasks.

**(5) Discourse-Aware**[9] (Cohan et al., 2018): An abstractive summarization model with a hierarchical encoder for discourse structure and an attentive discourse-aware decoder.

**(6) SciBERTSUM**[10] (Sefid & Giles, 2022): An extractive summarization model which enhances BERTSUM by incorporating section embeddings and a sparse attention mechanism.

**(7) ExtSum-LG+RdLoss**[11] (Xiao & Carenini, 2020): An extractive summarization model that incorporates s a redundancy loss term during the sentence scoring phase to reduce sentence redundancy.

**(8) BigBird-Pegasus**[12] (Zaheer et al., 2020): A sparse attention-based model that extends the Transformer-based Pegasus to handle longer text.

**(9) GPT-4**[13]: A generative language model developed by OpenAI, recognized as the most powerful generative LLM to date.

### 4.3 Implementation details

**Structural function recognition:** Ma et al. (2022) highlight the critical influence

---

of title and content positioning in scientific articles on structural recognition. Inspired by them, we conducted controlled experiments to investigate the impact of different combinations of content on the SFR task:

**(1) Titles:** The title is a carefully chosen phrase by the author that reflects the theme and structure of the chapter.

**(2) Full Chapter Text:** Complete chapter contents providing comprehensive information for understanding the organizational structure.

**(3) Head and Tail of Chapter Contents:** The head and tail portions of each chapter contain the author's overall introduction and summary, which are crucial for structural recognition.

We employed the SciBERT within the PyTorch[14] for training the SFR model, with the weights sourced from *'scibert_scivocab_uncased'* on Hugging Face. The model was trained on a GPU with A5000-24G, with the dataset split into training, validation, and testing sets in a ratio of 8:1:1. In a preliminary experiment, we use {1e-4, 1e-5, 2e-5} as learning rate value and {4, 8, 16, 32} as batch size value. We obtain the best results with a learning rate of 1e-5 and a batch size of 16. Other hyper-parameters were fine-tuned according to the performance on validation datasets. In the end, the model was trained for 10 epochs with a batch size of 16 for training and 8 for validation. We used Adam optimizer for acceleration and applied a warm-up training strategy with 100 steps to expedite convergence. For the other models, we choose the appropriate parameters based on the original paper to give them optimal performance.

**Abstractive summarization:** The PEGASUS, BART, BigBird and T5-base and Longformer were sourced from Hugging Face and all in base-uncased mode, while the other models were obtained from the original repositories as provided in the corresponding papers. We set the model-generated summaries to a length range of 50 to 300 words. For models such as BART, where the input length is exceeded model's capacity to process the full text, we use a divide-and-conquer approach to generate short summary for each section, which are then concatenated to create the final summary. For

---

Longformer, we input both the chapter structure information and the content of the sections together for abstractive summarization. We conducted a preliminary with different values to balance the trade-off between summary quality and computational cost. For the n-gram repeat limit, we tested values ranging from 2 to 4 to control repetition in the generated summaries, and obtain the best results at 2. In beam search, we experimented with the number of beams from 3 to 7 and achieved the best results with 5, while other parameters were set to default values. To account for the varying contributions of each section to the overall summary, we followed the work of Ermakova et al.(2018) and Li & Xu (2023), assigning 30%, 25%, 30%, and 15% weights to the *Background*, *Methods*, *Results*, and *Conclusions* sections, respectively. As a result, the lengths of the *Background*, *Methods*, *Results*, and *Conclusions* sections were no more than 90, 75, 90, and 45 words, respectively. For GPT-4 (*gpt-4-0613*), we used the API provided by OpenAI, and input the source article with the prompt "Summarize the following scientific paper no more than 300 words", and the other parameters in the API are set to default values.

### 4.4 Evaluation Metrics

**Evaluation Metrics on the SFR task**. The precision, recall, and $F_1$-score were selected to evaluate classification performance. The formulas used are as follows:

$$= \frac{n}{\phantom{x}} \qquad (7)$$

$$\text{Recall} = \frac{\phantom{x}}{t \qquad \text{Number of chapters that}} \qquad (8)$$

$$F_1\_\text{Score} = 2 \times \frac{\text{Precision} \times \text{Recall}}{\text{Precision} + \text{Recall}} \qquad (9)$$

Where $C$ is a particular category of the chapter. To evaluate the overall performance across all categories, we utilize the Sklearn package to compute the macro-average of precision, recall, and F1-score as follows:

$$Macro\_P = \frac{1}{n}\sum_{i=1}^{n} P_i \qquad （10）$$

$$Macro\_R = \frac{1}{n}\sum_{i=1}^{n} R_i \qquad （11）$$

$$Macro\_F_1 = 2\times \frac{Macro\_P \times Macro\_R}{Macro\_P + Macro\_R} \qquad （12）$$

Where $n$ is the number of classification categories, $P_i$ and $R_i$ represent the precision and recall of a specific category of chapters.

**Evaluation Metrics on the AS task**. We employ Recall-Oriented Understudy for Gisting Evaluation (ROUGE-N) to computer the lexical overlap between generated summaries and the golden truth human-written abstracts. The formula of ROUGE-N are as follows:

$$R_N^{recall} = \frac{\sum_{S \in reference} \sum_{gram_N in S} Count_{match}(gram_N)}{\sum_{S \in reference} \sum_{gram_N in S} Count(gram_N)} \qquad （13）$$

$$R_N^{recall} = \frac{\sum_{S \in reference} \sum_{gram_N in S} Count_{match}(gram_N)}{\sum_{S \in reference} \sum_{gram_N in S} Count(gram_N)} \qquad （14）$$

$$R_N^{F_1} = \frac{2 \cdot R_N^{precision} \cdot R_N^{recall}}{R_N^{precision} + R_N^{recall}} \qquad （15）$$

where $S \in reference$ denotes a sentence or document in the reference set (which is the golden truth), $gram_N$ represents the sequence of $N$ consecutive words within the reference sentence, $Count_{match}(gram_N)$ indicates the number of times the N-gram appears in both the prediction and the reference, and $Count(gram_N)$ is the total occurrence of the N-gram in the reference text. In this paper, we utilize the office package based on python to compute ROUGE-1, ROUGE-2, and ROUGE-L. According to the official ROUGE evaluation script[15], all reported ROUGE scores have a 95% confidence interval in this paper.

Additionally, we selected the GEM-score(Ermakova et al., 2018) to evaluate the comprehensiveness of the generated summary, which is calculated as the sum of the weights of section classes $w(sc)$ that appear in both the summary and the source article. However, while GEM measures the coverage of important sections, it does not account for the efficiency or conciseness of the summary. To address this, we integrated the compression rate (CR) to balance between content coverage and summary length. The combined metric, $GEM_{CR}$ is defined as follows:

$$GEM_{CR} = \frac{\sum_{sc \in (ASC \cap FTSC)} w(sc)}{\sum_{sc \in FTSC} w(sc)} \bullet Norm(\frac{L_{source\ article}}{L_{generated\ summary}}) \qquad (16)$$

where *FTSC* and *ASC* represent the section classes in the source article and in the generated summary, respectively. $w(sc)$ denotes the importance weight of a section. $L_{source\ article}$ and $L_{generated\ summary}$ represent the lengths of the source article and the generated summary, respectively. *Norm* refers the normalization function.

---

[15] https://pypi.org/project/pyrouge/

**Implementation Details of GEM$_{CR}$ Metric.** The GEM$_{CR}$ metric needs to identify whether the summary includes a specific section structure. To accomplish this, we used Stanford CoreNLP[16] to segment the sentences in the generated summary, and then applied sentence-BERT [17] (Reimers & Gurevych, 2019) to compute the cosine similarity between each sentence and the sections of the source article. The section category with the highest similarity score is then assigned as the corresponding category for each sentence.

## 5. Results

In this section, we first report the results of the SFR task, followed by the results of the AR task. Additionally, to better evaluate the model's performance, we also include the results of human evaluation and ablation experiments. Finally, we provide a case study for detailed sample analysis.

### 5.1 Results on the SFR task

Experimental results for the SFR performance are presented in Table 4. The traditional deep learning method Bi-LSTM performs the worst on the PubMed and arXiv datasets with Macro_F1 values of 89.41% and 85.40%, respectively. The pre-trained models outperform the traditional deep learning methods, with SciBERT achieving the best results on both PubMed and arXiv datasets, with Macro_F1 values of 91.79% and 88.42%, respectively, particularly demonstrating a significant performance lead on the arXiv dataset. Additionally, SciBERT outperforms BERT with

---

[16] https://stanfordnlp.github.io/CoreNLP/
[17] https://github.com/UKPLab/sentence-transformers

improvements of 1.06% on the PubMed dataset and 0.82% on the arXiv dataset in terms of Macro_F1. This highlights the effectiveness of domain-specific pre-training in

Table 4. The overall performance of different models on the SFR task.

| Model | | PubMed | | | arXiv | | |
| --- | --- | --- | --- | --- | --- | --- | --- |
| | | Macro_P(%) | Macro_R(%) | Macro_F$_1$(%) | Macro_P(%) | Macro_R(%) | Macro_F$_1$(%) |
| BiLSTM | − | 89.97 | 88.85 | 89.41 | 86.21 | 84.62 | 85.40 |
| | CI | [89.65, 90.29] | [88.54, 89.20] | [89.05, 89.74] | [85.92, 86.58] | [84.29,84.81] | [85.10, 85.71] |
| BiLSTM-ATT | − | 90.12 | 89.39 | 89.75 | 86.74 | 85.29 | 86.01 |
| | CI | [89.71, 90.55] | [89.04, 89.72] | [89.40, 90.18] | [86.39, 87.13] | [84.96, 85.87] | [85.69, 86.52] |
| BERT | − | 91.35 | 90.11 | 90.73 | 88.19 | 87.02 | 87.60 |
| | CI | [90.02, 91.70] | [89.83, 90.53] | [90.42, 91.14] | [87.89, 88.57] | [86.77, 87.37] | [87.31, 87.89] |
| RoBERTa | − | 91.72 | 90.67 | 91.20 | 88.33 | 87.17 | 87.74 |
| | CI | [91.41, 91.92] | [90.35, 91.07] | [90.82, 91.65] | [88.01, 88.62] | [86.90, 87.48] | [87.43, 88.06] |
| T5-base | - | 92.21 | **91.24** | 91.72 | 87.90 | 86.61 | 87.25 |
| | CI | [91.80, 92.63] | [90.85, 91.67] | [91.39, 92.01] | [87.55, 88.24] | [86.22, 87.04] | [86.91, 87.65] |
| **SciBERT** | − | **92.38** | 91.21 | **91.79** | **89.01**[†] | **87.84**[†] | **88.42**[†] |
| | CI | [92.01, 92.82] | [90.88, 91.64] | [91.42, 92.06] | [88.77, 89.42] | [87.51, 88.20] | [88.10, 88.73] |

*where the best results are in ***bold*** and CI represents the 95% confidence interval. A † symbol indicates a significant difference compared to other models.

Table 5．The impact of different chapter compositions on the SFR task (Macro_$F_1$ %).

| Composition | | PubMed | | | arXiv | | |
|---|---|---|---|---|---|---|---|
| | | T5-base | RoBERTa | SciBERT | T5-base | RoBERTa | SciBERT |
| Chapter title | – | 89.74 | 89.71 | **90.59**[†] | 84.96 | 85.21 | **86.77**[†] |
| | CI | [89.46, 89.98] | [89.48, 90.08] | [90.22, 90.94] | [84.73, 85.23] | [84.90, 85.65] | [86.39, 87.21] |
| Chapter text | – | 89.92 | 88.94 | **90.47**[†] | 85.38 | 85.46 | **86.24**[†] |
| | CI | [89.60, 90.17] | [88.71, 89.16] | [90.21, 90.72] | [85.10, 85.76] | [85.15, 85.86] | [85.87, 86.57] |
| Title+Chapter text | – | 91.43 | 91.02 | **91.68** | 85.92 | 85.97 | **86.93**[†] |
| | CI | [91.07, 91.70] | [90.61, 91.44] | [91.30, 91.97] | [85.45, 86.42] | [85.56, 86.38] | [86.51, 87.42] |
| Title+25%(head+tail) | – | 90.81 | 90.53 | **91.18** | 86.11 | 86.28 | **87.48**[†] |
| | CI | [90.38, 91.05] | [90.05, 90.73] | [90.73, 91.53] | [85.84, 86.40] | [85.91, 86.64] | [87.04, 87.70] |
| Title+50%(head+tail) | - | 91.34 | 90.76 | **91.42** | 87.25 | 87.74 | **88.42**[†][*] |
| | CI | [90.92, 91.76] | [90.31, 91.21] | [90.15, 91.78] | [86.91, 87.65] | [87.43, 88.06] | [88.10, 88.73] |
| Title+75%(head+tail) | – | 91.72 | 91.20 | **91.79** | 86.87 | 87.29 | **88.15**[†] |
| | CI | [91.39, 92.01] | [90.82, 91.65] | [91.42, 92.06] | [86.41, 87.36] | [86.88, 87.76] | [87.74, 88.63] |
| #Average | – | 90.83 | 90.36 | **91.19** | 86.08 | 86.33 | **87.34**[†] |
| | CI | [90.47, 91.11] | [90.00, 90.71] | [90.67, 91.50] | [85.74, 86.47] | [85.97, 86.73] | [86.94, 87.71] |

*where the best results are in ***bold***, and CI represents the 95% confidence interval. A † symbol indicates a significant difference compared to other models. A ∗ symbol indicates a composition that significantly outperforms other compositions.

enhancing classification performance on SFR task. Moreover, it is evident that all models demonstrate better performance on the PubMed dataset compared to the arXiv dataset. This difference can be attributed to several factors: First, articles on the PubMed dataset represent finalized research works that have undergone review and standardization processes, whereas the arXiv dataset contains unpublished preprints from a wide range of disciplines. The complexity and diverse structure of arXiv articles make it difficult for the models to accurately capture the necessary patterns in the SFR task. Second, arXiv articles are generally longer than PubMed articles, which poses challenges for pretrained models constrained by token limits, typically around 512 tokens. As a result, important nuances and details beyond this limit may not be fully captured.

Table 5 presents the impact of different chapter composition on the SFR task, focusing on the performance of three closely related models: T5, RoBERTa, and SciBERT. Results indicate that the SciBERT model, employing only the chapter title provided by authors yields a Macro_$F_1$ score of 90.59% for the PubMed dataset and 86.77% for the arXiv dataset, which surpass those obtained when using the full chapter content as input. This can be attributed to the conciseness of titles, which are used by the authors to convey the core content, while full-text content tends to include more noisy text. Furthermore, when combining titles with chapter content, the Macro_$F_1$ score improves as the proportion of content increases. The optimal head-to-tail ratio for the PubMed dataset is 75%, though the 50% ratio yields similar results with no significant difference observed, as the confidence intervals for both ratios overlap, indicating that the differences are not pronounced. However, the 75% ratio shows a slightly higher average score and more favorable upper and lower bounds in the confidence interval. This suggests that while both ratios perform comparably, the 75% ratio may provide slightly better stability and robustness in the results. For the arXiv dataset, the optimal ratio is 50%. In addition, SciBERT achieved the best Macro_$F_1$ across all compositions and average scores on both datasets, with a notable advantage on the arXiv dataset. It's important to note that the average chapter length in the arXiv dataset exceeds 790 words, surpassing SciBERT's 512-token input limit. Similarly, sections like *Results* and *Conclusions* in the PubMed dataset also exceed this limit. Therefore, filtering out important information, such as the beginning and end of chapters, is beneficial for reducing input length while enhancing the performance of the SFR task.

Table 6. Training Efficiency of SciBERT, RoBERTa and T5-base on the SFR Task

| Model | #Size | Training time/epoch | | Optimal epoch | | Macro_$F_1$(%) | |
|---|---|---|---|---|---|---|---|
| | | PubMed | arXiv | PubMed | arXiv | PubMed | arXiv |
| RoBERTa | 125M | 92±10 min | 37±4min | 7 | 5 | 91.19 | 87.77 |
| T5-base | 220M | 147±16min | 58±7min | 9 | 7 | 91.72 | 87.23 |
| **SciBERT** | **110M** | **80 ±7 min** | **33±3min** | **6** | **3** | **91.76** | **88.46** |

Although SciBERT achieved notably better results on the arXiv dataset, the overlapping confidence intervals on the PubMed dataset indicate no statistically significant differences between SciBERT, T5-base, and RoBERTa in certain cases. In light of these results, we further investigated the training efficiency of these models to assess their practicality for large-scale tasks. As shown in Table 6, SciBERT, despite being the smallest model with only 110M parameters, consistently achieves the highest Macro_$F_1$ scores on both the PubMed (91.79%) and arXiv (88.42%) datasets. It outperforms larger models like T5-base (220M) and RoBERTa (125M), highlighting its robustness in handling diverse scientific domains. Moreover, SciBERT reached optimal performance with fewer training epochs and shorter training times. Specifically, SciBERT converged in 6 epochs on PubMed and 3 epochs on arXiv, compared to T5-base, which needed 9 and 7 epochs, and RoBERTa, which took 7 and 5 epochs, respectively. In terms of training time, SciBERT was also more efficient, averaging 1 hour and 20 minutes per epoch on PubMed and 33 minutes per epoch on the arXiv dataset. These results highlight SciBERT's significant advantages in training efficiency, computational demands, and energy consumption, which is beneficial for reducing hardware dependency during local deployment. Therefore, in order to achieve a balanced performance and operational cost, SciBERT was selected as the backbone for the SFR task.

Table 7. ROUGE-1/2/L scores of different models on the PubMed and arXiv datasets.

| Models | PubMed | | | | arXiv | | | |
|---|---|---|---|---|---|---|---|---|
| | ROUGE-1 | ROUGE-2 | ROUGE-L | $GEM_{CR}$ | ROUGE-1 | ROUGE-2 | ROUGE-L | $GEM_{CR}$ |
| BERTSUM | 34.39 | 13.24 | 30.90 | 0.42 | 31.98 | 10.02 | 27.75 | 0.25 |
| BART | 37.44 | 15.39 | 32.71 | 0.51 | 34.87 | 12.82 | 29.83 | 0.37 |
| PEGASUS | 36.94 | 15.05 | 31.81 | 0.48 | 33.36 | 11.18 | 28.94 | 0.32 |
| T5-base | 37.75 | 14.92 | 32.65 | 0.50 | 33.75 | 11.76 | 29.01 | 0.33 |
| Discourse-Aware | 38.93 | 15.37 | 35.21 | 0.59 | 35.80 | 11.05 | 31.80 | 0.42 |
| SciBERTSUM | 45.13 | 19.03 | 40.80 | 0.65 | 44.05 | 15.67 | 39.15 | 0.64 |
| BigBird-Pegasus | 42.81 | 18.71 | 39.23 | 0.63 | 42.47 | 17.95 | 37.29 | 0.59 |
| ExtSum-LG+RdLoss | 45.30 | **20.42** | 40.95 | 0.66 | 44.01 | 17.79 | 39.09 | 0.62 |
| GPT-4 | 30.02 | 09.17 | 27.62 | 0.74 | 29.96 | 09.13 | 27.01 | 0.71 |
| Longformer | **45.38** | 19.27 | **40.97** | **0.83** | **44.51** | **18.26** | **39.64** | **0.78** |

## 5.2 Results on the AS task

Experimental results for the AS performance are presented in Table 7. Results show that the Longformer achieves the highest scores in ROUGE-1, ROUGE-2, ROUGE-L and $GEM_{CR}$, with scores of 44.51, 18.26, 39.64 and 0.69 for the arXiv dataset, and 45.38, 19.27, 40.97 and 0.71 for the PubMed dataset, respectively. This indicates that the summaries generated by Longformer are closer to human-written abstracts lexical and comprehensive aspects. In contrast, models such as PEGASUS, and T5 struggle with long scientific papers due to input length limitations. Although divide-and-conquer strategies have been employed to address these limitations, a significant portion of the arXiv dataset remains challenging for these models to capture text patterns. As a result, their performance is less robust compared to Longformer and BigBird-Pegasus. The ExtSum-LG+RdLoss model performed better than the PLMs such as BART, PEGASUS, and T5-base. This is because ExtSum-LG+RdLoss is an extractive model that forms summaries by directly extracting sentences from the original article, which more accurately reflects the lexical similarity of the article compared to generative models.

Additionally, despite GPT-4's outstanding performance on various general NLP tasks, there is still considerable room for improvement in scientific paper summarization when evaluated with ROUGE scores. One possible explanation is that we only use a simple prompt to guide GPT-4 in generating summarization results, with no restrictions on content or format. As a result, the summary generated by GPT-4 is more random and diverse, which may not align well with the human-written abstract at the lexical level (Wang et al., 2023; Yang et al., 2023). From a comprehensive perspective, GPT-4 achieved the second-best $GEM_{CR}$ score, indicating that despite lexical differences from human-written summaries, it still effectively captures and conveys key information during summarization.

Figure 4 shows the length distribution of generated summaries by BART (Figure 4a and 4b), Longformer (Figure 4c and 4d) and GPT-4(Figure 4e and 4f) across the PubMed and arXiv datasets. As mentioned in Section 3.2, scientific papers in the arXiv dataset have longer abstracts compared to those in the PubMed dataset. When the

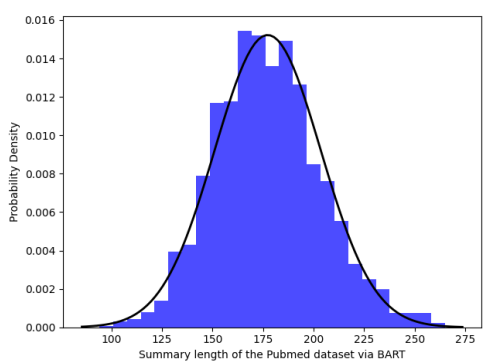

(a) Length distribution of generated summaries by BART on the PubMed dataset

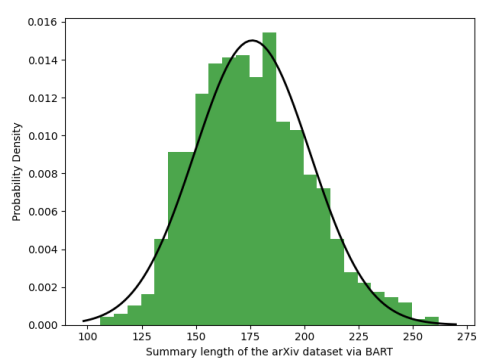

(b) Length distribution of generated summaries by BART on the arXiv dataset

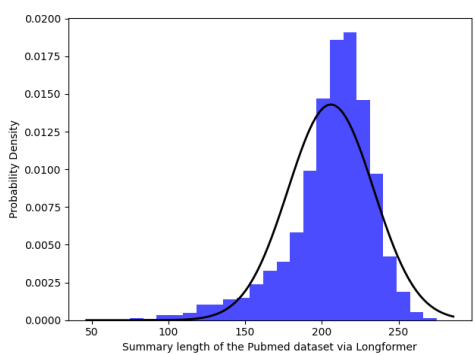

(c) Length distribution of generated summaries by Longformer on the PubMed dataset

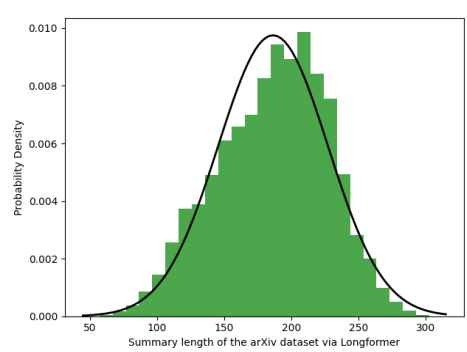

(d) Length distribution of generated summaries by Longformer on the arXiv dataset

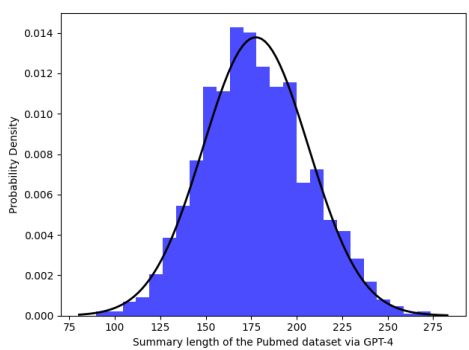

(e) Length distribution of generated summaries by GPT-4 on the PubMed dataset

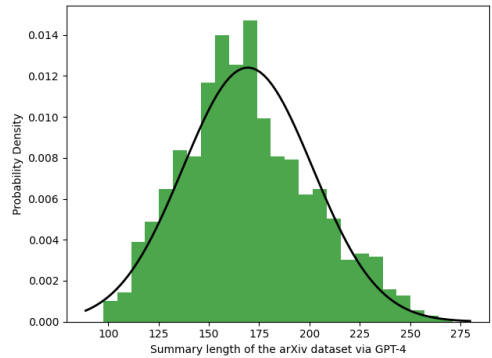

(f) Length distribution of generated summaries by GPT-4 on the arXiv dataset

Figure 4. The distribution of summary lengths generated by different models, with the x-axis representing the length of the generated summaries and the y-axis representing the probability density.

model's output length is limited, there are minimal differences observed in the range of summaries lengths generated by the PLMs. However, the distribution of these summary lengths differs. For instance, both BART and GPT-4 consistently generate summaries clustered between 150-190, whereas Longformer's summaries are centered around 200 to 220 words in both datasets, closely aligning with the average abstract lengths in the original data. This alignment likely contributes to Longformer's superior performance on both datasets.

Table 8. The criteria for human-oriented informativeness score.

| Informativeness score | Criteria |
| --- | --- |
| 1 | The summary includes very little or none of the key information from the human-written abstract. |
| 2 | The summary includes some information but misses the key points from the human-written abstract. |
| 3 | The summary includes around half of the key points from the human-written abstract but lacks critical details. |
| 4 | The summary includes most of the key points from the human-written abstract, with only minor details missing. |
| 5 | The summary includes almost all key points from the human-written abstract with strong detail and accuracy. |

Table 9. The criteria for human-oriented coherence score.

| Coherence score | Criteria |
| --- | --- |
| 1 | The summary lacks coherence, with sentences that are disconnected and lack logical flow. |
| 2 | The summary has limited coherence, with noticeable issues in logical flow or organization. |
| 3 | The summary is generally coherent, though it contains minor breaks in flow or organization. |
| 4 | The summary is mostly coherent, with clear logical flow and only occasional minor disruptions in structure or clarity. |
| 5 | The summary is fully coherent, with seamless logical flow and structure, making it easy to understand. |

Table 10. The criteria for human-oriented readability score.

| Readability score | Criteria |
| --- | --- |
| 1 | The language is disjointed, with significant grammar or word usage errors, and difficult to read. |
| 2 | The language contains noticeable grammar or word usage errors, but the overall meaning can still be read. |
| 3 | The language flows generally smooth but contains a few grammatical or word usage errors, it remains readable. |
| 4 | The language reads mostly smooth, with minor grammar or word usage errors, but overall reads well. |
| 5 | The language flows very smooth, with no grammar or word usage errors, and is easy to read. |

### 5.3 Human evaluation vs LLM evaluation

#### *5.3.1 Results on the Human evaluation*

In our comparison using ROUGE scores, we observed that GPT-4, although considered the most advanced generative model, had the lowest scores. However, since ROUGE measures lexical overlap between generated summaries and human-written abstracts, it's important to note that a low ROUGE score doesn't indicate a poor summary quality(Fu et al., 2024; Liu et al., 2023). Therefore, to provide a more comprehensive evaluation, we conducted a human evaluation for the generated summaries from the informativeness, coherence, and readability perspectives.

In the first step, we randomly selected a total of 100 summaries from the PubMed and arXiv datasets. The summaries generated by the baseline model, as well as the associated golden truth human-written abstracts, are provided. To mitigate potential biases, the models are labeled with numbers rather than their actual names. Then, three second-year doctoral students majoring in information science were invited to compare the generated summaries with the corresponding golden truth abstracts and rate them based on three aspects: (1) How informative is the generated summary? (3) How coherent is the generated summary? (2) How readability is the generated summary? Each aspect of the summaries is scored on a scale ranging from 1 to 5, with 1 is the worst and 5 indicating the best. The criteria of informativeness, coherence and readability are shown in Table 8, Table 9, and Table 10, respectively.

Table 11．The human evaluation results for summaries generated by different models.

| Models | Informativeness | Coherence | Readability | #Average |
|---|---|---|---|---|
| BERTSUM | 2.38 | 2.16 | 2.45 | 2.33 |
| BART | 3.01 | 2.89 | 2.91 | 2.94 |
| PEGASUS | 2.82 | 2.75 | 3.04 | 2.87 |
| T5-base | 2.94 | 3.22 | 2.97 | 3.04 |
| Discourse-Aware | 3.23 | 3.39 | 3.13 | 3.25 |
| SciBERTSUM | 3.59 | 3.67 | 3.38 | 3.55 |
| BigBird-Pegasus | 3.28 | 3.56 | 3.59 | 3.48 |
| ExtSum-LG+RdLoss | 3.52 | 3.65 | 3.72 | 3.63 |
| Longformer | 3.61 | 3.76 | 3.85 | 3.74 |
| GPT-4 | **3.64** | **3.94** | **4.17** | **3.91** |

As shown in Table 11, despite GPT-4's lower ROUGE scores, it achieved the highest overall performance across all three aspects of human evaluation. Notably, the summaries generated by GPT-4 exhibited significantly high readability, consistent with the findings of (Lozić & Štular, 2023). As the most advanced generative model to date, GPT-4 is designed to generate content that align with human preferences, which is beneficial for improving the fluency of summaries. Additionally, the higher informativeness and coherence score from GPT-4 indicate its strength in capturing key information and maintaining logical connections. Finally, Longformer, the model selected in our study, achieved the second-highest performance in human evaluation, demonstrating its robustness under different evaluation scenarios.

### 5.3.2 Results on the LLM evaluation

To balance the potential biases in human evaluation due to individual differences in knowledge, we also using LLM to evaluate the summaries from the same criteria as the human evaluation. For instance, we selected G-EVAL (Liu et al., 2023) metric, which is specifically designed to use GPT-4 to assess the quality of natural language generation outputs through the chain-of-thought (CoT). The process of LLM evaluation is shown in Figure 5. In this experiment, we first guided the LLM to rate the human-written abstracts based on the criteria, and then we conducted a comparative evaluation between the generated summaries and the golden truth abstracts.

---

**_(1) Task Introduction_**

You will be given a generated summary and a Human-written summary for a scientific paper. Your task is to rate the generated summary on three aspects: informativeness, readability, coherence. The criteria for each aspect are as follows:

- Criteria for informativeness:
- Criteria for readability:
- Criteria for coherence:

**_(2) Evaluation Steps_**

1. Read the Human-written summary carefully and assess its key points, clarity, and logical flow as a reference.

2. Read the generated summary and compare it to the Human-written summary.

3. Assign a score for each aspect on a scale of 1 to 5, where 1 is the lowest and 5 is the highest based on the Evaluation Criteria.

4. Your output should only be the score, without additional comments or explanations.

*Human-written summary:*
*Generated Summary:*
- *Informativeness:*
- *Coherence:*
- *Readability:*

---

Figure 5．The evaluation process using G-EVAL for generated summaries.

Table 12．The results of the LLM evaluation of the generated summaries from different models.

| Models | Informativeness | Coherence | Readability | #Average |
|---|---|---|---|---|
| Human-written | 4.51 | 4.48 | 4.60 | 4.53 |
| BERTSUM | 2.81 | 2.46 | 2.71 | 2.66 |
| BART | 3.37 | 3.06 | 3.18 | 3.28 |
| PEGASUS | 3.32 | 3.29 | 3.29 | 3.22 |
| T5-base | 3.31 | 3.54 | 3.20 | 3.35 |
| Discourse-Aware | 3.73 | 3.75 | 3.38 | 3.62 |
| SciBERTSUM | 3.83 | 3.98 | 3.61 | 3.80 |
| BigBird-Pegasus | 3.64 | 3.95 | 3.87 | 3.82 |
| ExtSum-LG+RdLoss | 3.84 | 4.02 | 4.09 | 3.98 |
| Longformer | 3.98 | 4.14 | 4.15 | 4.09 |
| GPT-4 | **4.20** | **4.27** | **4.42** | **4.30** |

As shown in Table 12, the human-written abstracts received the highest ratings in the LLM evaluation. The scores of all three aspects increased compared to the human evaluation, but there is still room for improvement to match the standard of human-written abstracts. Excluding from the human-written abstracts, the summaries generated by GPT-4 still received the highest score. Although studies have found that LLM-based evaluations tend to provide overly positive feedback (Fu et al., 2024), summaries generated by Longformer still achieved the second-highest performance under the same evaluation criteria and models, demonstrating its robustness across different evaluation scenarios. Additionally, this observation also the biases in traditional summary evaluation metrics, such as ROUGE, which primarily evaluate summary quality based on lexical overlap with golden truth texts, and may not fully capture the strengths of generative models like GPT-4 in summarization tasks.

## 5.4 Ablation study

We conducted ablation experiments to further validate the impact of structure information in scientific paper summarization. In the first step, we removed all section headings and retained only the article title and full-text content. Then, based on this, we separately added the structure provided by the author, the structure obtained through direct keyword matching from the mapping file, and the structure recognized through SFR task in our proposed method.

Table 13．The abstractive summarization performance of the Longformer w/o structure information.

| Model | PubMed | | | |
| --- | --- | --- | --- | --- |
| | ROUGE-1 | ROUGE-2 | ROUGE-L | GEM$_{CR}$ |
| Ours | **45.38** | 19.27 | **40.97** | **0.83** |
| *w/* title + full-text | 43.72 | 17.75 | 38.48 | 0.55 |
| *w/* title + full-text + author-provided structure | 44.08 | 19.25 | 39.64 | 0.70 |
| *w/* title + full-text + keywords mapping structure | 44.12 | **19.30** | 39.73 | 0.74 |
| Model | arXiv | | | |
| | ROUGE-1 | ROUGE-2 | ROUGE-L | GEM$_{CR}$ |
| Ours | **44.51** | **18.26** | **39.64** | **0.78** |
| *w/* title + full-text | 41.76 | 15.62 | 37.18 | 0.59 |
| *w/* title + full-text + author-provided structure | 43.28 | 18.18 | 38.29 | 0.64 |
| *w/* title + full-text + keywords mapping structure | 42.77 | 18.14 | 37.96 | 0.61 |

As shown in Table 13, when only the article title and full-text content were used, the performance of Longformer dropped substantially across both datasets. Specifically, ROUGE 1/2/L and $GEM_{CR}$ scores dropped by 1.66, 1.52, 2.49, and 0.28 on the PubMed dataset, respectively. Similarly, the scores decreased by 2.75, 2.64, 2.46, and 0.19 on the arXiv dataset, respectively. With the addition of the author-provided structure, both ROUGE and $GEM_{CR}$ scores showed significant improvement. We speculate that this improvement arises because most human-written abstracts include key components such as *Background*, *Methods*, *Results*, and *Conclusions*. Therefore, incorporating chapter structure information helps the summaries better reflect the structural elements of the paper, thus enhancing alignment with human-written abstracts. Moreover, a notable trend observed is that the method using direct keyword mapping via a mapping file slightly outperformed the author-provided structure but still fell short of the performance achieved by the SFR task. This is because the mapping file was derived from the biomedical domain, making it well-suited to the PubMed dataset. In contrast, for the arXiv dataset, which predominantly focuses on computer science and physics, the keyword mapping proves less effective due to differences in structure and terminology across these fields.

Overall, the observed conclusions demonstrate that, firstly, normalizing section titles to focus on four main sections helps mitigate the structural uncertainty introduced by author-provided titles. Secondly, training a classifier on a large corpus for the SFR task provides a more stable and effective method for structural function recognition compared to the direct use of keyword mapping, further validating the positive contributions of different components in our proposed framework.

### 5.5 Case study

Figure 6 presents an example of a scientific paper from the PubMed dataset. The first section displays the original abstract, while the second, third, and fourth sections report the output summaries of the BART, Longformer, and GPT-4 models, respectively. It is observed that Longformer notably includes details on the analysis of "*serum blood urea nitrogen (bun), creatinin, na+, k+, cl+, ca+, p+, and creatininity clearance was*

| | |
|---|---|
| Original | Radiocontrast administration is an important cause of acute renal failure. in this study , compared the plasma creatinine levels with spot urine il-18 levels following radiocontrast administration. twenty patients (11 males, 9 females ) underwent radiocontrast diagnostic and therapeutic - enhanced examinations. the rin mehrin risk score was low (5). the radiocontrast agents used were 623 mg / ml iopromid (1.5 ml / kg ), and 100 ml of 650 mg / ml meglumine diatrizoate as three - way oral and rectal contrast material for abdominal computed tomography ( ct ) scans . serum blood urea nitrogen , creatinin , na, k; cl, ca, p , creatinine clearance , and spot urine il-18 levels were analyzed before and repeated at 24 , 48 , and 72 h after radiocontrast administration. six and 24-h urinary il-18 levels were measured with a human il-18 elisa kit following radiocontrast administration. an increase in plasma creatinine 24 and 48 h following radiocontrast administration was observed compared with precontrast values, but it was not statistically significant ( p=0.052 and p=0.285 , respectively). a statistically significant increase in il-18 levels was observed at 6 and 24 h , compared with precontrast values ( p=0.048 and p=0.028 , respectively). a tendency for postcontrast 24-h urinary il-18 levels to increase was observed compared with 6 h , but the increase was not statistically significant (p=0.808). our results show that plasma creatinine starts to increase at 24th hour ; however , spot urineil-18 levels go up at 6th hour following radiocontrast administration implying urine il-18 to be an earlier parameter for kidney injury. |
| BART | B1:Radiocontrast-induced nephropathy (rin) can lead to acute renal failure (arf). Arf increases treatment cost due to sepsis, hemorrhage, respiratory failure, and a long hospitalization.<br>M1:twenty patients (11 males and 9 females) underwent diagnostic and therapeutic contrast - enhanced examinations at the department of internal medicine from january 2009 to march 2009. The study was approved by the institute ethics committee and written consent was obtained from the selected patients based on a low mehran risk score (5) serum creatinine levels increased after radiocontrast administration, although the result was not statistically significant.<br>R3:a slight increase in creat in 48h was observed, but they fell to pre Contrast values at 72 h. spot urine il-18 levels in patients were slightly increased.<br>C3:The most common definition of rin is plasma creatinine levels of 0.5 mg/dl or higher 72h after contrast administration or 25% higher than the basal Plasma creatinine level. The first 24 h remains unclear in patients with acute renal injury, but il-18 levels start to increase within 4-6 h.<br>ROUGE-1: 0.4387   ROUGE-2: 0.1571   ROUGE-L: 0.4238 |
| Longformer | B1:Radiocontrast-induced nephropathy (rin) is an important cause of hospital-acquired acute renal failure (arf) and is responsible for 12% of cases.<br>B19:recent studies have reported that urine levels of il-18(a pro-inflammatory cytokine), kidney injury molecule-1, and neutrophil gelatinase - associated lipocalin (ngal) levels are important for early detection of rin.<br>B21:in this study, we aimed to compare the plasma creatinine levels with spot urine il-8 levels following intravenous iopromid (623 mg/ml,1.5 ml/kg;ultravist 300) administration.<br>M1:twenty patients (11 males and 9 females)underwent diagnostic and therapeutic contrast-enhanced examinations at the department of internal medicine from january 2009 to march 2009.<br>M7:a precontrast-enhanced examination of serum blood urea nitrogen (bun), creatinine, na+, k+, cl+, ca+, p+ and creatininy clearance was analyzed and they were repeated at 24, 48, and 72h following contrast administration. spot urine il8 levels were measured with a human il8 elisa kit before and 6 and 24 h after intravenous iopromid administration.<br>M30:the statistical analysis was performed with the ncss pass 2007 and 2008 statistical software (kaysville,ut, usa)<br>R3:a slight increase in the level of spot urinary il8 was observed at 24 and 48h following radioclast administration, but it was not statistically significant (p=0.052 and p= 0.285, respectively).<br>ROUGE-1: 0.5779   ROUGE-2: 0.2851   ROUGE-L: 0.5519 |
| GPT-4 | B1: Radiocontrast-induced nephropathy (RIN) can cause acute renal failure (ARF), leading to dialysis therapy. ARF increases treatment costs due to complications such as sepsis, hemorrhage, respiratory failure, and lengthy hospital stays. RIN is responsible for 12% of hospital-acquired ARF cases. Possible mechanisms for RIN include renal medullary hypoxia and the toxic effects of contrast agents on renal tubules. Risk factors for RIN include renal insuff<br>M1:Twenty patients (11 males and 9 females) underwent contrast -enhanced examinations at the Department of Internal Medicine from January to March 2009.<br>?:The study had ethical approval and written consent from the patients. Pre-and post-contrast examinations were conducted to analyze various blood and urine parameters. Specific contrast materials were used for the examinations.<br>R1:After receiving radiocontrast, serum creatinine levels increased but were not statistically significant. However, there was a slight increase in creatinine levels at 48 hours, which returned to precontrast levels at 72 hours. Plasma creatinine levels also slightly increased at 24 and 48 hours, but again, it was not statistically significant.<br>R9:On the other hand, postcontrast urinary levels of il8 at 6 and 24 hours significantly increased compared to precontrast levels.<br>C4:There was a tendency for The common definition of renal injury (RIN) is a plasma creatinine level of 0.5 mg/dl or higher 72 hours after contrast administration or 25% higher than the baseline level. Plasma creatinine levels start to rise within 24 hours in 80% of RIN patients, peaking at 48-72 hours, and returning to normal after 2 weeks. The first 24 hours are unclear, but il-18 levels begin to increase within 4-6h.<br>ROUGE-1: 0.3377   ROUGE-2: 0.0744   ROUGE-L: 0.3245 |

Figure 6. A comparative case study between the generated summary and the golden truth abstracts from the source paper. (The source article is from the paper PMC3459524, doi:10.4103/0971-4065.98756). Letters to the left of each sentence indicate different sections, where B, M, R, and C stand for Background, Methods, Results, and Conclusion, and the numbers indicate the respective line positions within each section. The portions overlapping with the golden summary are highlighted in green.

*analyzed and they were repeated at 24, 48, and 72h following contrast administration.* *spot urine il8 levels*" and "*was observed at 24 and 48h following radioclast administration, but it was not statistically significant (p=0.052 and p= 0.285, respectively)*" which is the main method and the result of the source article. Additionally, Longformer captures longer continuous sentences and phrases, while GPT-4 focuses on generating summaries across the entire document but does not consistently align with the human-written abstract. Furthermore, all generative models tend to extract sentences from the opening parts of different sections in a scientific paper, which aligns with findings in the SFR task that the beginning and ending parts of sections often contain more important information.

## 6. Discussion

In this section, we summarize the main findings of this paper and discuss their implications. Then, we highlight some limitations of our work and suggest areas for further improvement.

### 6.1 Implications

Automated summarization of scientific papers is important for managing large volumes of literature and enhancing research efficiency in the field. In this paper, we propose a two-stage framework for abstractive summarization of scientific papers. In the first stage, we automatically identify the main sections (e.g. *Background*, *Method*, *Result*, and *Conclusion*) of scientific papers that are most relevant to the human-written abstract, which helps improve the comprehensiveness of the summary while reducing computational complexity. In this task, the SciBERT mode demonstrates superior performance compared to other deep learning methods. Additionally, we observe that the beginning and end of each chapter contain more important information, which contributes to improving the structural function recognition in scientific papers.

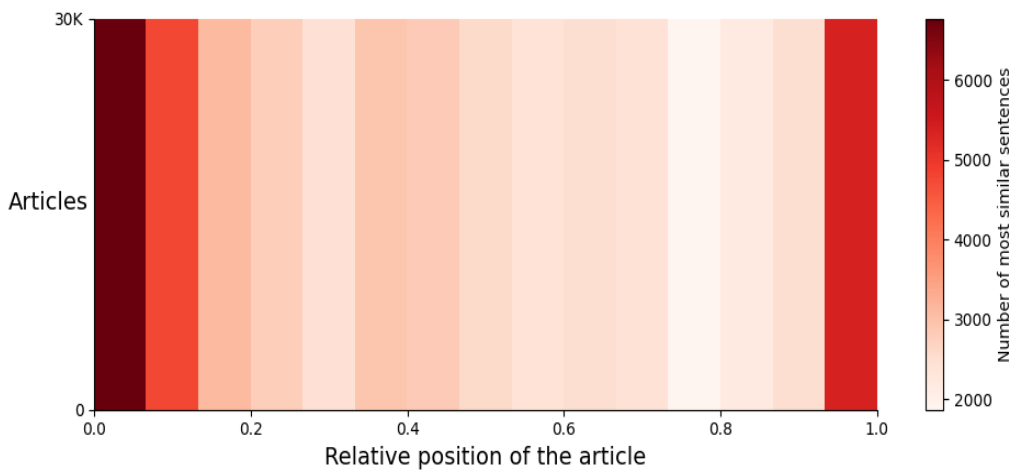

(a) Positional similarities between full-text and abstract sentences in the PubMed dataset

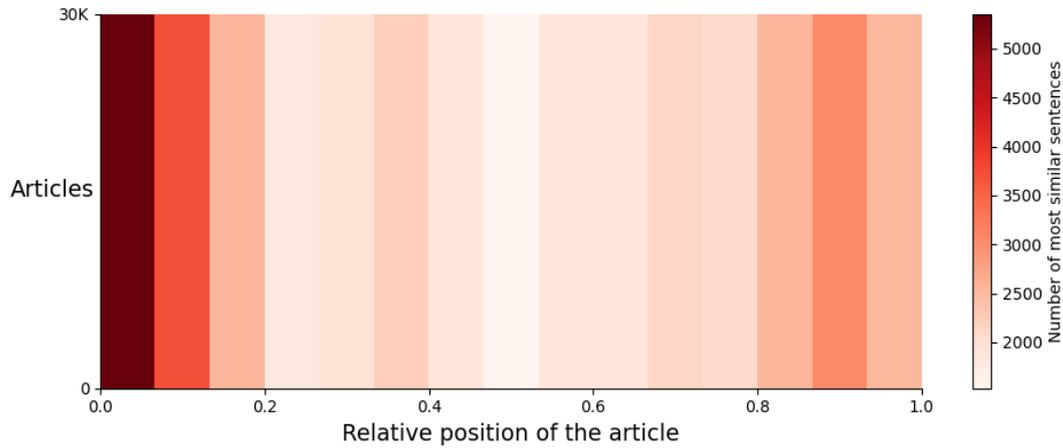

(b) Positional similarities between full-text and abstract sentences in the arXiv Dataset

Figure 7. The Distribution of sentence positions in arXiv and PubMed articles corresponding to abstracts.

Based on this finding, we further analyzed the distribution of similar sentences between abstracts and full texts using cosine similarity, as illustrated in Figure 7. The results highlight a clear preference for the beginning and end of the articles, which are more conducive to summarization. Furthermore, we conduct a comparative analysis of mainstream PLMs in abstractive summarization. The results show that Longformer outperform other baseline models. In contrast, GPT-4 exhibits the lowest ROUGE scores when compared to human-written abstracts, yet achieves high performance in both human evaluations and LLM evaluations. A case study reveals that GPT-4 frequently employs connectives, pronouns, and excessive elaboration, contributing to its lower ROUGE scores. However, its strong performance in human and LLM assessments highlights GPT-4's strengths in coherence and readability. These findings highlight the bias and limitations of using traditional summary evaluation metrics, such as ROUGE and BLEU, in assess generative models like GPT-4.

Finally, our research holds notable implications for fields that heavily rely on literature reviews. For example, in the computer science and biology domains, rapid technological advancements have resulted in a high volume of new publications daily. Summarizing these papers is crucial for improving research efficiency in these areas. In particular, the advancement of open access policies providing convenient to freely download these papers in PDF format. Moreover, recent developments in efficient PDF parsing tools, such as GPTpdf[18], LlamaParse[19] and PymuPDF[20], have further enhanced the ability to process scientific documents. Therefore, in this context, our work can serve as a foundation for future automated summarization from scientific papers, and further contributing to research on review generation.

## 6.2 Limitations

This paper has several limitations. Firstly, our proposed two-stage framework for abstractive summarization of scientific articles, which includes two sequential tasks: structural function recognition and abstractive summarization, may introduce potential errors in task transmission. Secondly, while we applied PLMs like BART, PEGASUS, T5, and Longformer for abstractive summarization, we did not fine-tune them specifically to the arXiv and PubMed datasets. Fine-tuning these models on domain-specific datasets could help enhance their performance for specific tasks (Liu, 2019). Lastly, due to equipment limitations, our proposed two-stage model did not utilize larger and more recent models, so its performance could be further improved in the future.

Another limitation is that our model, while designed to work effectively with papers following the IMRaD structure, may struggle with non-IMRaD articles. Specifically, the modeling of scientific paper structures has always been one of the research hotspots. Although the IMRaD format is widely adopted in the natural sciences, models such as the Harmsze(Harmsze, 2000) also hold an important place in the structure of scientific papers. Additionally, for papers in the social sciences, there is greater structure flexibility due to the use of varied theoretical frameworks and a wide range of research methods(Mohajan, 2018). Therefore, IMRaD format is not universally applicable across all disciplines, and exploring more generalizable structures for organizing scientific papers will be a focus in the future.

---

[18] https://github.com/CosmosShadow/gptpdf/
[19] https://github.com/run-llama/llama_parse
[20] https://github.com/pymupdf/PyMuPDF

# 7. Conclusion and future work

In this paper, we propose a novel two-stage framework for abstractive summarization of scientific papers, which is based on automatic structural function recognition to generate comprehensive summaries. To achieve this, we first normalized the chapter titles of numerous scientific papers and reconstructed a structural function recognition corpus. We then developed a long text classification model by fine-tuning a pre-trained model on this corpus, and achieved improved performance by leveraging key information from the beginning and end of chapters. Next, this model was used to automatically recognize important structural functions such as *Background*, *Method*, *Results*, and *Conclusion*, where were input into a generative model to produce a comprehensive and balanced summary. Additionally, in view of the fact that most traditional summarization evaluation metrics only reflect lexical overlap with human-written summaries, we further designed new evaluation criteria to assess the generated summaries from both human and LLM perspectives.

In the experiments, our model outperformed mainstream PLMs on two publicly available scientific paper summarization datasets (PubMed, arXiv), and the summaries generated by our model were more comprehensive. We also demonstrated the positive impact of chapter structure information on improving the quality and completeness of scientific paper summarization.

We hope our work provides valuable insights into the application of abstractive summarization for scientific papers. Meanwhile, we notice that the proposed two-stage framework could be optimized by exploring alternative backbones. In our future research, more advanced LLM will be investigated for structure information extraction and abstractive summarization to further enhance performance.

**ACKNOWLEDGMENTS**

This paper was supported by the National Natural Science Foundation of China (Grant No.72074113, 72374103).